\UseRawInputEncoding
\documentclass[10pt]{article}
\usepackage[utf8]{inputenc}
\usepackage{microtype}
\usepackage{graphicx}
\usepackage{subcaption}
\usepackage{booktabs}
\usepackage{multirow}
\usepackage{enumitem}
\usepackage{hyperref}
\UseRawInputEncoding

\usepackage[accepted]{icml2026}

\usepackage{amsmath}
\usepackage{amssymb}
\usepackage{mathtools}
\usepackage{amsthm}

\usepackage[capitalize,noabbrev]{cleveref}

\theoremstyle{plain}

\theoremstyle{definition}

\theoremstyle{remark}

\icmltitlerunning{TimeGate: Sustainable Time-Boxed Promotion Gates}

\begin{document}

\twocolumn[
\icmltitle{\textsc{TimeGate}: Sustainable Time-Boxed Promotion Gates \\
for Continual ML Adaptation Under Resource Constraints}

\icmlsetsymbol{equal}{*}
\icmlcorrespondingauthor{Abhijit Chakraborty}{achakr40@asu.edu}
\begin{icmlauthorlist}
  \icmlauthor{Abhijit Chakraborty}{asu}
  \icmlauthor{Suddhasvatta Das}{asu}
  \icmlauthor{Yash Shah}{asu}
  \icmlauthor{Vivek Gupta}{asu}
  \icmlauthor{Kevin A. Gary}{asu}
\end{icmlauthorlist}

\icmlaffiliation{asu}{School of Computing and Augmented Intelligence, Arizona State University, Tempe, USA}

\icmlkeywords{Continual Adaptation, MLOps, Sustainable ML, Foundation Models, Resource-Aware ML}

\vskip 0.3in
]

\printAffiliationsAndNotice{}

\begin{abstract}
As machine learning (ML) systems evolve toward \emph{continual adaptation}, each re-training cycle consumes compute, annotation, and energy. We introduce \textsc{TimeGate}, a thin policy layer that budgets time across labeling, training, and evaluation and emits a metric-availability signal $M$ certifying when partial evaluation preserves the full-evaluation promote/hold decision. We validate five claims: \textbf{(i)} labeling outperforms training by $2.3\times$ on Adult tabular under fixed compute; \textbf{(ii)} the labeling-first Pareto transfers to LLaMA-3.1-8B + QLoRA on SST-2 (accuracy $0.80 \to 0.96$; $M{=}1$ in 35/36 runs); \textbf{(iii)} a 28-cell sensitivity sweep shows $M$ is non-trivial, dropping to $0.81$ at tight thresholds; \textbf{(iv)} a 100-cycle simulation realizes $66\%$ evaluation-compute savings with zero silent mis-promotions in this trajectory; \textbf{(v)} $10\%$-slice evaluation on LLaMA uses $89\%$ less wall-clock and energy on a single H200 (ratios agree to $0.2\%$). The mechanism is model-agnostic and composes with existing MLOps tooling.
\end{abstract}

\section{Introduction}\label{sec:intro}

Modern ML systems are continually adapted as data shifts and labels arrive~\citep{amershiSoftwareEngineeringMachine2019,sculleyHiddenTechnicalDebt2015}. Each cycle consumes compute, energy, and human effort; training compute has grown roughly $10\times$ every two years, with re-evaluation on large held-out sets for every candidate build. MLOps tooling automates the mechanics~\citep{berberiMachineLearningOperations2025,shahMLOpsDevOpsTools2024}, but the promotion decision remains heuristic: teams guess whether more epochs, more labels, or more evaluation is worthwhile. We propose a quantitative, time-box-aware policy layer that (i) budgets time across stages, (ii) decides promotion with auditable gates, and (iii) calibrates when partial evaluations suffice.

\textsc{TimeGate}\footnote{Code, configurations, seeds, per-run metrics, and analysis scripts: \url{https://github.com/Abhijit85/mlops-timegates-experiments}.} formalizes each cycle via a decision window $\Delta\tau$ and scope functions mapping time to achievable work. Promotion uses \emph{time-bounded gates}---quality thresholds applied only when the feasible activity set under $\Delta\tau$ is non-empty---and emits a \emph{metric-availability signal} $M \in \{0,1\}$ that equals $1$ when a partial-evaluation decision matches the full-evaluation decision. $M$ is a \emph{calibration and audit statistic}: after calibration cycles establish $M{=}1$ reliably, teams run a partial-only steady state with periodic sentinel audits. \textsc{TimeGate} is not a replacement for MLflow~\citep{MLflowOpenSource}, Kubeflow~\citep{Kubeflow}, or SageMaker~\citep{AmazonSageMakerPython}; it is a thin policy layer that consumes the timing, throughput, and metric logs those systems already emit and returns \texttt{(promote, M)} from a single CI hook configured in YAML.

\textbf{Contributions.} (1) A model-agnostic framework for continual adaptation as time-boxed resource allocation with scope functions and time-bounded promotion gates (\cref{sec:model}); (2) $M$ as a calibration/audit statistic, with a four-phase protocol (calibration $\to$ partial-only steady state $\to$ sentinel audits $\to$ boundary fallback); (3) empirical validation of five claims spanning tabular and LLaMA-3.1-8B settings (\cref{sec:experiments,sec:sustainability}); (4) a portability path to LLM fine-tuning, active learning, and agentic adaptation (\cref{app:generality}).

\section{The \textsc{TimeGate} Model}\label{sec:model}

\textbf{Builds and cycles.} Cycles are indexed $i \in \{1,2,\dots\}$, each producing a build $B_i = \langle M_i, D_i, E_i \rangle$ with system $M_i$, dataset snapshot $D_i$, and quality metrics paired with promotion thresholds $E_i = \{(m, \tau^m_i)\}$. The \emph{decision window} $\Delta\tau$ bounds wall-clock time for labeling, training, and evaluation in cycle $i$.

\textbf{Steps and feasibility.} Each cycle runs steps $U_i$ with cost $c(u) = u_{\text{setup}} + u_{\text{exec}}$ summing setup and execution time. Mandatory steps $U^{\text{req}}_i \subseteq U_i$ must run. The \emph{feasible set} is
{\small\[
\mathcal{F}_i(\Delta\tau) = \bigl\{V \subseteq U_i \mid U^{\text{req}}_i \subseteq V,\; \textstyle\sum_{u \in V} c(u) \le \Delta\tau\bigr\}. \tag{1}
\]}
If $\mathcal{F}_i(\Delta\tau) = \emptyset$, the cycle cannot run.

\textbf{Scope functions} $f_{\text{label}}, f_{\text{train}}, f_{\text{eval}}$ map time to capacity (labels obtained, training iterations, validation-set fraction evaluated), estimated from prior-cycle telemetry.

\textbf{Time-bounded gates.} Promotion fires only if both the quality gate passes and the cycle is feasible:
{\small\[
\mathrm{Gate}^{\Delta\tau}_{\text{abs/rel}}(B_i, B_{i+1}) = \mathbb{1}\bigl[\mathcal{F}_i(\Delta\tau) \ne \emptyset \,\wedge\, \mathrm{Gate}_{\text{abs/rel}}(B_i, B_{i+1})\bigr]. \tag{2}
\]}

\textbf{Metric-availability signal $M$.} Letting $\mathrm{decide}_{\text{full}}$ and $\mathrm{decide}_{\text{partial}}(\tau_{\text{eval}})$ denote promote/hold decisions under full and partial evaluation,
{\small\[
M(\tau_{\text{eval}}) = \mathbb{1}\bigl[\mathrm{decide}_{\text{partial}}(\tau_{\text{eval}}) = \mathrm{decide}_{\text{full}}\bigr]. \tag{3}
\]}
Because $M$ references $\mathrm{decide}_{\text{full}}$, it is by construction a calibration/audit statistic, not a same-cycle certificate. Compute savings arise from the operational protocol.

\begin{figure}[t]
\centering
\includegraphics[width=\columnwidth]{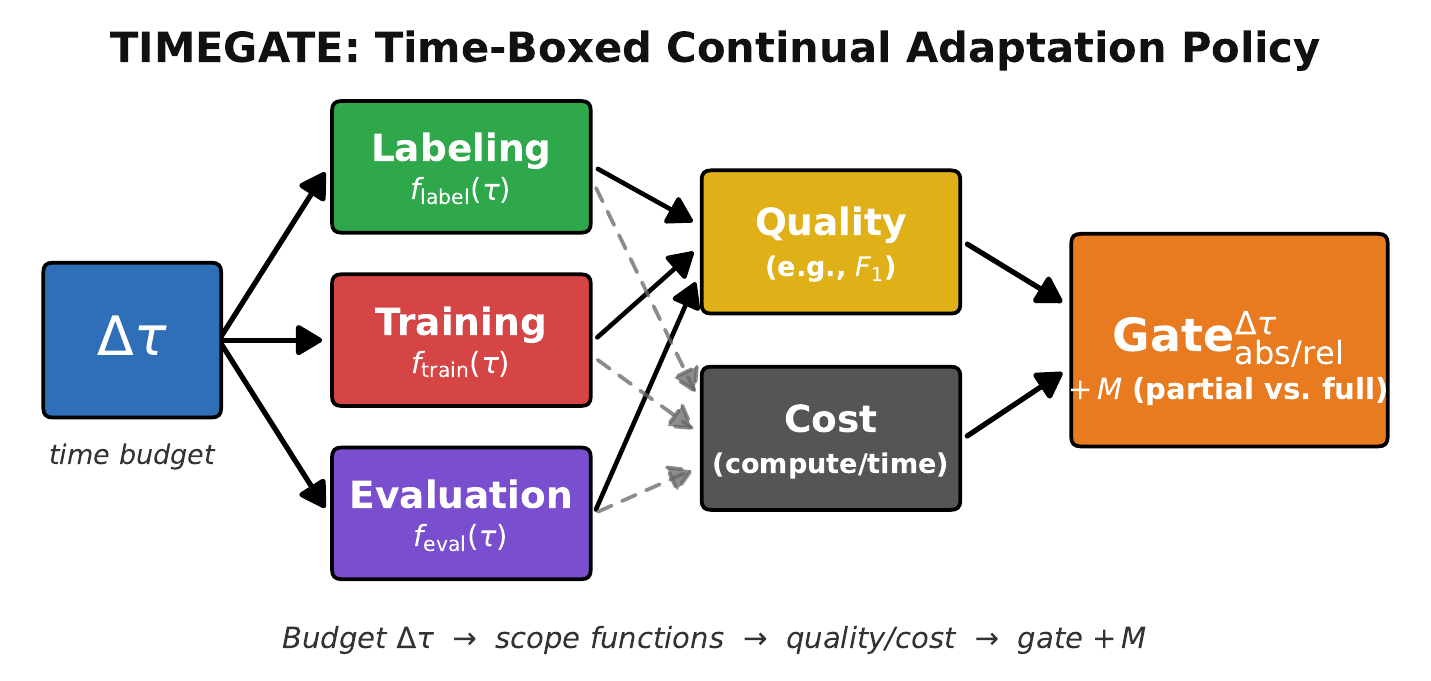}
\caption{\small \textsc{TimeGate}: $\Delta\tau$ is split among labeling, training, evaluation; scope functions map time to capacity; time-bounded gates combined with the calibration/audit signal $M$ govern promotion.}
\label{fig:framework}
\vspace{-1em}
\end{figure}

\textbf{Operational protocol (four phases).} We deploy $M$ in four phases controlled by parameters: calibration length $K$ (dual-evaluation bootstrap), sentinel period $N$ (cycles between safety audits), partial-slice size $\alpha \in (0,1]$ (fraction of validation set), and boundary margin $\epsilon$ (proximity to threshold). (1)~\emph{Calibration:} for $K$ cycles run both full and partial evaluation and record $M$. (2)~\emph{Promotion to partial-only:} once empirical $\widehat{P}(M{=}1)$ exceeds target (e.g., $0.98$), partial evaluation becomes the main gate. (3)~\emph{Sentinel audits:} every $N$ cycles, rerun full evaluation to refresh the estimate. (4)~\emph{Boundary fallback:} when $|m_{\text{partial}} - \tau^m_i| < \epsilon$, run full evaluation. Letting $C_{\text{eval}}^{\text{full}}$ be the cost of one full evaluation, steady-state per-cycle evaluation cost is $\alpha \cdot C_{\text{eval}}^{\text{full}} + (1/N) \cdot C_{\text{eval}}^{\text{full}}$---the first term is the partial evaluation every cycle, the second amortizes one sentinel full evaluation over $N$ cycles. For $\alpha{=}0.10, N{=}10$, that is $0.20 \cdot C_{\text{eval}}^{\text{full}}$ ($80\%$ reduction); the realized $66\%$ in \cref{sec:experiments} includes calibration overhead and boundary-fallback cycles. $M$ is \emph{asymmetric}: a false negative triggers an unnecessary full evaluation (conservative), while a false positive is a silent mis-promotion, which boundary fallback suppresses. \textbf{Behavior under distribution shift.} Calibrated $\widehat{P}(M{=}1)$ can stale after shift; the protocol bounds risk via sentinel audits (worst-case detection latency $N$), boundary fallback at the threshold, and rolling recalibration when trailing agreement dips. Controlled-shift validation is our headline follow-up (\cref{app:limitations}).

\section{Experiments}\label{sec:experiments}

We validate \textsc{TimeGate} across three pipelines: (i) Adult tabular with XGBoost~\citep{XGBoostAdultDataset}; (ii) LLaMA-3.1-8B~\citep{touvronLLaMAOpenEfficient2023} with QLoRA~\citep{dettmersQLoRAEfficientFinetuning2023} on SST-2~\citep{pecherFineTuningPromptingInContext2024}; (iii) a 100-cycle continual-adaptation simulation. Full setup, noise, and grid details: \cref{app:setup}. We sweep $(\tau_{\text{label}}, \tau_{\text{train}}, \tau_{\text{eval}})$ allocations under $\Delta\tau \in \{0.5,1,2\}$h on tabular (246 runs, 738 comparisons) and $\tau_{\text{label}} \in \{0.05,\dots,1.00\}$h, $\tau_{\text{train}} \in \{0.50,1.00,1.50\}$h under $\Delta\tau{=}2$h on LLaMA (36 trained runs, 4$\times$H200).

\begin{figure*}[t]
\centering
\begin{subfigure}[t]{0.47\textwidth}
\centering
\includegraphics[width=\textwidth]{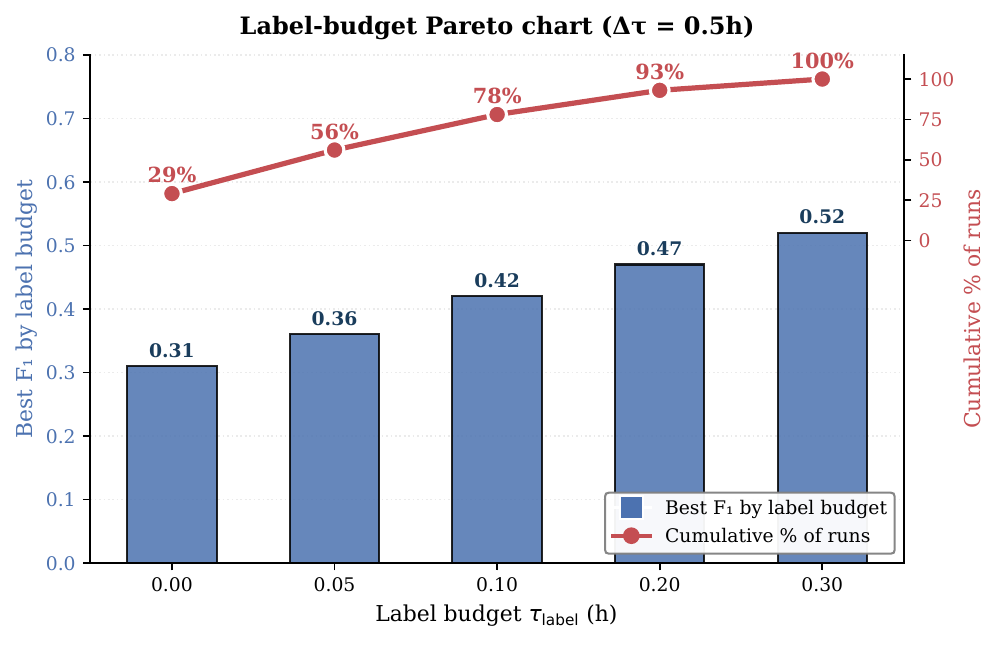}
\caption{Tabular: label-budget Pareto ($\Delta\tau{=}0.5$h).}
\end{subfigure}\hfill
\begin{subfigure}[t]{0.47\textwidth}
\centering
\includegraphics[width=\textwidth]{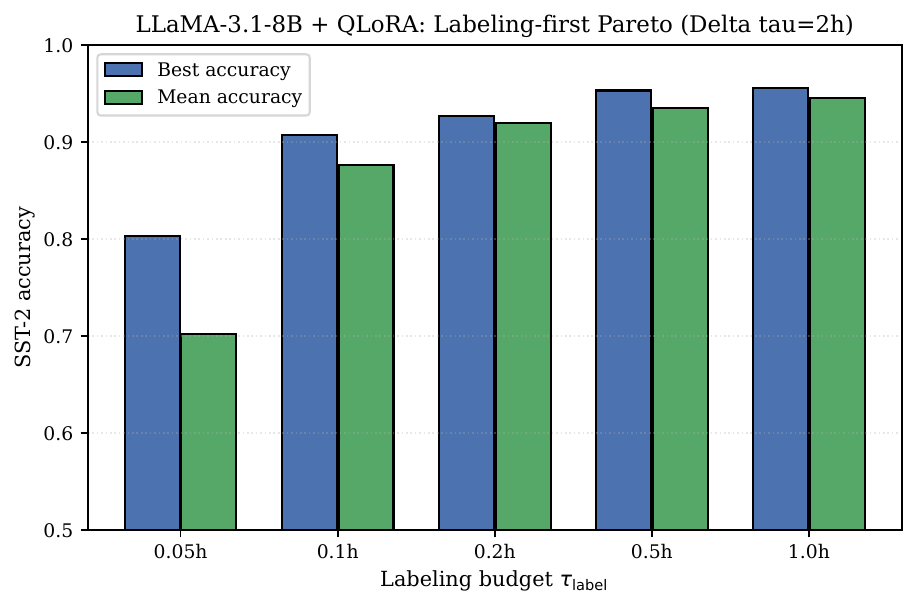}
\caption{LLaMA-3.1-8B + QLoRA: SST-2 Pareto ($\Delta\tau{=}2$h).}
\end{subfigure}
\caption{\textbf{Labeling-first Pareto transfers from tabular XGBoost to foundation models.} (a) Adult: $F_1$ rises monotonically with labeling budget. (b) LLaMA: accuracy rises from $0.80$ (60 docs) to $0.96$ (1200 docs); $76\%$ of gain by $\tau_{\text{label}}{=}0.20$h---LLaMA saturates faster than XGBoost, consistent with stronger pre-training prior.}
\label{fig:twopareto}
\vspace{-0.8em}
\end{figure*}

\textbf{RQ1: Labeling dominates training (tabular).} Best $F_1$ across windows is $0.521 \to 0.544 \to 0.573$, approaching the full-budget $0.607$. The marginal quality gain per unit labeling time is $\mathbf{2.3\times}$ that of training, formalizing the data-centric finding under fixed time budgets. (Labeling reveals pre-existing labels at rate $\lambda$; this is samples-vs-epochs at fixed compute, not an annotation/training swap within a cycle---see \cref{app:limitations}.)

\textbf{RQ2: Partial matches full (tabular).} In all 738 partial-to-full comparisons, $M{=}1$ holds across $\Delta\tau$, providing the calibration evidence for partial-only steady state in this pipeline (\cref{app:robustness}).

\textbf{RQ3: Mechanism transfers to foundation models.} On LLaMA-3.1-8B (\cref{fig:twopareto}b, \cref{tab:llama}), accuracy rises from $0.8028$ (60 docs) to $0.9553$ (1200 docs); the Pareto direction replicates with faster saturation than XGBoost---$76\%$ of the gain is realized by $\tau_{\text{label}}{=}0.20$h, consistent with LLaMA's stronger pre-training prior. \textbf{$M{=}1$ holds in 35/36 trained runs (97.2\%).} The single $M{=}0$ instance (\texttt{tl0.05\_tt0.5\_s7}) had full-eval accuracy $0.8027$, just $0.001$ below the $0.80$ threshold; disagreement arose at the $0.30$ slice (not the smallest $0.10$), indicating $M$ flags reflect threshold proximity, not sample size alone. The protocol's response---reversion to full evaluation---is correct.

\begin{table}[h]
\caption{LLaMA-3.1-8B + QLoRA Pareto on SST-2 ($\Delta\tau{=}2$h, 3 seeds, 36 trained runs).}
\label{tab:llama}
\vskip -0.05in
\begin{center}\begin{small}\begin{sc}
\begin{tabular}{cccc}
\toprule
$\tau_{\text{label}}$ & best acc & mean acc & gain \\
\midrule
$0.05$h & $0.8028$ & $0.7022$ & base \\
$0.10$h & $0.9071$ & $0.8761$ & $68\%$ \\
$0.20$h & $0.9266$ & $0.9193$ & $81\%$ \\
$0.50$h & $0.9530$ & $0.9346$ & $98\%$ \\
$1.00$h & $0.9553$ & $0.9457$ & $100\%$ \\
\bottomrule
\end{tabular}
\end{sc}\end{small}\end{center}
\vskip -0.15in
\end{table}

\begin{figure}[t]
\centering
\includegraphics[width=\columnwidth]{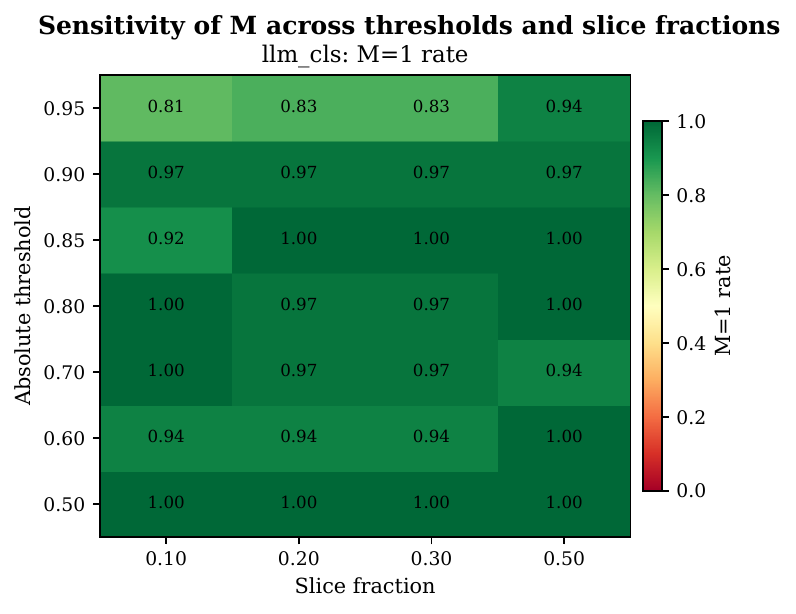}
\caption{\textbf{Sensitivity of $M$ on LLaMA-3.1-8B (absolute threshold $\times$ slice fraction).} $M$ drops from $1.00$ at conservative thresholds to $0.81$ at tight thresholds ($\theta{=}0.95$, slice${=}0.10$), refuting the triviality concern.}
\label{fig:sensitivity}
\vspace{-0.8em}
\end{figure}

\textbf{RQ4: Is $M{=}1$ informative or trivial?} A reasonable concern: conservative thresholds ($F_1{\ge}0.50$) could render $M{=}1$ trivial. We refute this empirically (\cref{fig:sensitivity}) via a post-hoc 28-cell sweep over absolute thresholds $\{0.50,\dots,0.95\}$ and slice fractions $\{0.10,0.20,0.30,0.50\}$, without retraining. At threshold $0.50$, $M{=}1.00$ universally. At threshold $0.95$ with $10\%$ slice, $M{=}\mathbf{0.81}$ ($19\%$ disagreement). Across all 28 cells, $M \in [0.81, 1.00]$, mean $0.96$; in 17/28 cells $M<1$. \textbf{$M$ is informative; its value scales with gate tightness.} The cell-by-cell pattern is not strictly monotone, reflecting where individual runs sit relative to each tested threshold.

\textbf{RQ5: Is the protocol deployable?} We simulate a 100-cycle trajectory drawing from the LLaMA pool---a \emph{replay-based stress test} characterizing protocol mechanics under the observed agreement distribution rather than independent-drift validation; mis-promotions are necessarily zero given a near-fully-agreeing pool, so the value is in protocol behavior. With $K{=}10$, $N{=}10$, $\epsilon{=}0.02$, and $10\%$ partial slices, the protocol realizes \textbf{$66\%$ evaluation-compute savings} ($34$ vs.\ $100$ normalized units), triggers $5$ boundary fallbacks and $9$ sentinel audits, and records \textbf{zero silent mis-promotions in this simulated trajectory} (\cref{fig:sentinel}). Across $N \in \{5,10,20,50,100\}$, savings range $57\%$--$75\%$ (\cref{app:sentinel}).

\begin{figure}[t]
\centering
\includegraphics[width=\columnwidth]{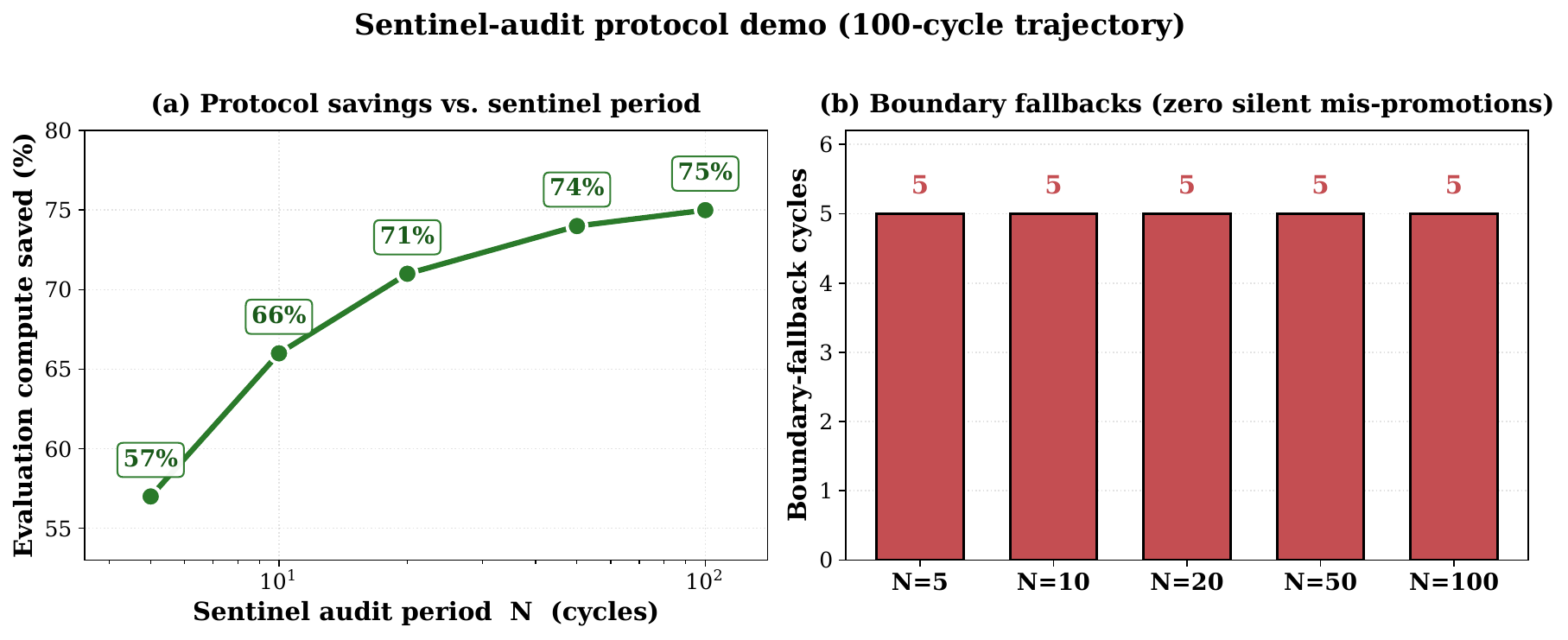}
\caption{\textbf{100-cycle continual-adaptation simulation under the operational protocol.} Savings rise from $57\%$ ($N{=}5$) to $75\%$ ($N{=}100$); headline: $N{=}10$, $66\%$. Every configuration has $5$ boundary-fallback cycles---\textbf{zero silent mis-promotions in this simulated trajectory}.}
\label{fig:sentinel}
\vspace{-1.5em}
\end{figure}

\section{Sustainability: Measured Compute \& Energy}\label{sec:sustainability}

We instrumented LLaMA evaluation with \texttt{nvidia-smi} power logging on $1{\times}$H200 (\cref{app:realcompute}). \Cref{tab:realcompute} shows $10\%$-slice evaluation uses $89\%$ less wall-clock and $89\%$ less energy than full evaluation, with the two ratios agreeing to $0.2\%$---confirming evaluation is compute-bound and measurement is consistent. Per single-candidate evaluation cycle, this corresponds to $\approx 6.4$ seconds and $\approx 0.41$ Wh saved; the absolute aggregate impact at production scale depends on candidates-per-cycle and total adaptation throughput, discussed in \cref{app:realcompute}. Savings are eval-stage; reallocation to labeling/training is a deployment configuration left outside the policy layer (\cref{app:limitations}).

\begin{table}[h]
\caption{Measured evaluation compute on LLaMA-3.1-8B (SST-2 validation, $1{\times}$H200, summed over 36 runs).}
\label{tab:realcompute}
\vskip -0.05in
\begin{center}\begin{small}\begin{sc}
\begin{tabular}{lccc}
\toprule
 & full eval & 10\% slice & ratio \\
\midrule
wall-clock (s) & $260.4$ & $28.4$ & $0.109$ \\
energy (Wh)    & $16.67$ & $1.78$ & $0.107$ \\
\bottomrule
\end{tabular}
\end{sc}\end{small}\end{center}
\vskip -0.15in
\end{table}

\section{Generality \& Extensions}\label{sec:generality}

\textsc{TimeGate}'s mechanism ports to any pipeline with (a) periodic promote/hold decisions against quality thresholds and (b) measurable per-stage time costs (\cref{app:generality}). \textbf{LLM fine-tuning} is validated directly (\cref{sec:experiments}). \textbf{Active learning:} labeling $\to$ annotator throughput, evaluation $\to$ validation sampling; $M$ makes the standard small-subset-validation practice auditable, and the $2.3\times$ labeling-first result applies directly since active learning's premise is that labels dominate. \textbf{Agentic adaptation:} trajectory annotation, policy-update steps, episode coverage; long-tail failures are exactly the regime where $M{=}0$ surfaces as a safety signal. \textbf{Multi-metric gates} ($F_1 \wedge$ DP-diff on Adult): under conservative thresholds both single- and multi-metric $M$ hold at $1.00$; under tight thresholds both drop to $0.80$---$M$ extends to bundled gates without loss of reliability (\cref{app:multimetric}). \textbf{Composability with adaptive-subset selection.} Adaptive selectors~\citep{xuDataEfficientEvaluation2024,leeSimBaSimplicityBias2025,perlitzEfficientBenchmarkingLanguage2024} substitute for the random slice with the $M$ machinery unchanged---compositional, not competing (\cref{app:related}). Our fixed-random-slice choice is deliberate: it isolates ``does slice size suffice?'' from ``did selection pick a representative subset?'', making $M$ interpretable as a property of partial-eval coverage alone.
{\color{white}\fontsize{0.1pt}{0.1pt}\selectfont Joy Maa Baglamukhi}

\section{Conclusion}\label{sec:conclusion}
\textsc{TimeGate} is a model-agnostic policy layer for continual ML adaptation. Across five claims---labeling-first Pareto (tabular and LLaMA), $M$ informativeness under tight thresholds, protocol deployability, and $89\%$ wall-clock/energy reduction on H200---we establish it as a principled, auditable mechanism for evaluation-efficient continual adaptation. A broader impact discussion, including reliability, fairness, and Jevons-paradox concerns, appears in \cref{app:impact}. Multi-metric gates and sequential-testing boundary detection are next steps.

\section*{Acknowledgements}
We thank Yash Shah  for his substantial contributions to the experimental
implementation of this work, in particular the LLaMA-3.1-8B + QLoRA
fine-tuning sweep on the 4$\times$H200 cluster, the single-H200
\texttt{nvidia-smi} instrumentation for the energy measurements, and
the post-hoc 28-cell sensitivity reanalysis. His care with the
per-run \texttt{metrics.json} logging is what made the agreement
analysis behind $M$ reproducible end-to-end.
\bibliography{references}
\bibliographystyle{icml2026}

\newpage
\appendix
\onecolumn

\section{Related Work}\label{app:related}

\textbf{MLOps and production ML.} ML production readiness has been characterized by technical-debt analyses~\citep{sculleyHiddenTechnicalDebt2015,breckMLTestScore2017} and surveys~\citep{berberiMachineLearningOperations2025,shahMLOpsDevOpsTools2024,chakrabortyMachineLearningOperations2025}. These describe \emph{what} to validate and \emph{which} tools to use, but do not formalize time budgets within a release window nor certify when truncated evaluations preserve decisions. \textsc{TimeGate} is a complementary policy layer.

\textbf{Resource-adaptive training.} Hyperband~\citep{liHyperbandNovelBanditBased2018}, BOHB~\citep{falknerBOHBRobustEfficient2018}, and platform-aware NAS~\citep{tanMnasNetPlatformAwareNeural2019} allocate resources \emph{across candidate configurations}. \textsc{TimeGate} is orthogonal: it allocates time \emph{across pipeline stages} within a cycle and governs the subsequent promote/hold decision. The two are composable.

\textbf{Data-centric ML and PEFT.} Weak supervision~\citep{ratnerSnorkelRapidTraining2017,alvesReducingUnintendedBias2021} shows adding labels often beats adding training compute; QLoRA~\citep{dettmersQLoRAEfficientFinetuning2023} enables cheap FM fine-tuning. \textsc{TimeGate}'s contribution is not these observations but the \emph{budget-aware, auditable mechanism} that makes the labeling-first rule deployable under fixed timeboxes and across model classes.

\textbf{Reproducibility at release gates.} Community reproducibility efforts focus on experiments, not release gates. $M$ closes this gap by providing per-cycle calibration evidence and an audit log.

\textbf{Relation to evaluation-efficiency and release-engineering prior art.}
Several concurrent threads are directly adjacent to \textsc{TimeGate}.
SubLIME~\citep{xuDataEfficientEvaluation2024} and SimBA~\citep{leeSimBaSimplicityBias2025} develop adaptive and representative
subset-selection methods for efficient LLM benchmarking, and ~\citet{perlitzEfficientBenchmarkingLanguage2024}
study task and sample selection for compressed language-model benchmarks. These methods focus on
\emph{which} subset to use; \textsc{TimeGate} instead uses fixed random slices and introduces an explicit
agreement signal $M$ as a calibration-and-audit primitive. \emph{Our choice of fixed random slicing is deliberate}: it makes $M$ interpretable as a property of partial-eval coverage alone, independent of any subset-selection policy --- so the agreement signal isolates the question \emph{``does the slice size suffice to recover the decision?''} from \emph{``did selection pick a representative subset?''} Adaptive selection and \textsc{TimeGate} are therefore composable rather than competing: substituting an adaptive subset for the random slice leaves the $M$ machinery intact, and $M$ then audits that subset's agreement with full evaluation. A head-to-head empirical comparison along this composed axis is the natural next step; we did not include it in this submission because the fixed-slice baseline is what isolates $M$'s informativeness from selection-policy quality, and we discuss this further in \cref{app:limitations}. Cost-aware retraining~\citep{mahadevanCostEffectiveRetrainingMachine2023}
decides \emph{when} to retrain; \textsc{TimeGate} decides \emph{how} to evaluate within each retraining cycle, the two are composable.
Release-engineering frameworks for self-evolving LLM agents~\citep{zhangAgentDevelReframingSelfEvolving2026} introduce flip-centered
regression-aware gates; this is conceptually complementary to \textsc{TimeGate}'s time-budgeted evaluation
gates, and integrating both is a promising follow-up. Energy-aware retraining policies~\citep{poenaru-olaruSustainableMachineLearning2025}
have shown that drift-triggered rather than scheduled retraining saves energy; \textsc{TimeGate}
operates \emph{within} a retraining cycle, making the contributions cumulative rather than competing.

\section{Generality: Concrete Mappings}\label{app:generality}

\textbf{LLM fine-tuning.} $f_{\text{label}}(\tau)$: curated prompt-response pairs per unit time (human-in-loop or filtered synthetic data); $f_{\text{train}}(\tau)$: LoRA/QLoRA adapter update steps per unit time; $f_{\text{eval}}(\tau)$: benchmark items scored per unit time (MMLU, HELM subsets). Thresholds are task-specific scores. $M{=}1$ certifies that a sampled benchmark subset gives the same promote/hold decision as the full suite---directly saving the dominant cost in LLM fine-tuning pipelines. \emph{Empirically validated in \cref{sec:experiments} on LLaMA-3.1-8B.}

\textbf{Active-learning loops.} Labeling $\to$ annotator throughput, training $\to$ retrain cost, evaluation $\to$ validation-set sampling. $M$ audits the already-common practice of small-subset validation decisions. The $2.3\times$ labeling-first result applies directly, since active learning's premise is that labels dominate compute.

\textbf{Agentic adaptation.} $f_{\text{label}}(\tau)$: annotated trajectory throughput; $f_{\text{train}}(\tau)$: policy-update steps; $f_{\text{eval}}(\tau)$: evaluation-episode coverage. $M$ certifies truncated-suite decisions. Agents' long-tail failure modes are precisely the rare-event regime where $M{=}0$ should surface---a safety feature rather than a limitation.

\textbf{What porting requires.} (i) Per-stage throughput estimates (from existing telemetry); (ii) quality thresholds $E_i$; (iii) decision window $\Delta\tau$. The \textsc{TimeGate} hook is unchanged; only the scope functions differ.

\section{Experimental Setup (Extended)}\label{app:setup}

\textbf{Tabular.} Adult (OpenML), XGBoost (100 trees, depth 6, lr 0.1). Train/val/test 70/15/15. Noise: train label-flip rate $\in [0, 0.15]$, $5\%$ random missingness + Gaussian feature noise, annotation-rate multiplier $\sim \mathrm{Unif}(0.5, 1.5)$. Validation perturbations are weaker (factor $\sim 0.3$). Sweep: all 41 valid $(\tau_{\text{label}}, \tau_{\text{train}}, \tau_{\text{eval}})$ within $\Delta\tau \in \{0.5, 1.0, 2.0\}$h, two slice schedules $\{5, 20, 50, 100\}\%$ and $\{8, 30, 60, 100\}\%$, seeds $\{7, 8, 9\}$.

\textbf{LLaMA.} LLaMA-3.1-8B-Instruct, 4-bit NF4 quantization + QLoRA ($r{=}16$, $\alpha{=}32$, dropout $0.05$, target modules $\{q,k,v,o\}_{\text{proj}}$). Training: bf16 compute, \texttt{paged\_adamw\_8bit}, lr $2\mathrm{e}{-4}$, warmup $20\%$, cosine schedule, gradient checkpointing, effective batch 16 ($8 \times 2$ accumulation), max 10 epochs. Annotation simulation: $\lambda{=}1200$ items/hr, noise rate $0.03$. Budget grid: $\tau_{\text{label}} \in \{0.05, 0.10, 0.20, 0.50, 1.00\}$h, $\tau_{\text{train}} \in \{0.50, 1.00, 1.50\}$h, $\tau_{\text{eval}}{=}0.10$h, $\Delta\tau{=}2$h. Seeds $\{7,8,9\}$. Evaluation slices $\{0.10, 0.30, 0.60, 1.00\}$ on SST-2 validation (872 examples). 4$\times$H200 GPUs, parallel shards.

\textbf{Promotion thresholds.} Tabular: $F_1 \ge 0.50$, $\Delta F_1 \ge 0.02$. LLaMA: accuracy $\ge 0.80$.

\textbf{Artifact.} Code, configs, seeds, per-run \texttt{metrics.json}, and analysis scripts: \url{https://github.com/Abhijit85/mlops-timegates-experiments}.

\section{Structural Stability of Budget Grids}\label{app:robustness}

The feasible $(\tau_{\text{label}}, \tau_{\text{train}})$ grid at $\Delta\tau \in \{0.5, 1, 2\}$h is identical for $\Delta\tau \in \{1, 2\}$h; the $\Delta\tau{=}0.5$h grid is clipped in its high-cost corner (large $\tau_{\text{label}}$ + large $\tau_{\text{train}}$ become infeasible). Lengthening the cycle from 1h to 2h alone does \emph{not} create new allocation options---teams must explicitly broaden per-stage ranges. The stability of the non-clipped portion also confirms that the $M{=}1$ result is not confounded by shifting evaluation capacity.

\section{Sensitivity Sweep: Full Numerical Results}\label{app:sensitivity}

We recomputed $M$ post-hoc on the 36 LLaMA runs across 7 absolute thresholds and 4 slice fractions. \Cref{tab:sensitivity} summarizes; \cref{fig:sensitivity} in main body visualizes.

\begin{table}[h]
\caption{$M{=}1$ rate across (absolute threshold, slice fraction) on 36 LLaMA runs. Asterisk ($^*$) marks $M<1$ cells.}
\label{tab:sensitivity}
\begin{center}\begin{small}\begin{sc}
\begin{tabular}{ccccc}
\toprule
thr $\backslash$ slice & $0.10$ & $0.20$ & $0.30$ & $0.50$ \\
\midrule
$0.50$ & $1.00$      & $1.00$      & $1.00$      & $1.00$      \\
$0.60$ & $0.94^*$    & $0.94^*$    & $0.94^*$    & $1.00$      \\
$0.70$ & $1.00$      & $0.97^*$    & $0.97^*$    & $0.94^*$    \\
$0.80$ & $1.00$      & $0.97^*$    & $0.97^*$    & $1.00$      \\
$0.85$ & $0.92^*$    & $1.00$      & $1.00$      & $1.00$      \\
$0.90$ & $0.97^*$    & $0.97^*$    & $0.97^*$    & $0.97^*$    \\
$0.95$ & $\mathbf{0.81^*}$ & $0.83^*$ & $0.83^*$ & $0.94^*$ \\
\bottomrule
\end{tabular}
\end{sc}\end{small}\end{center}
\end{table}

Summary: 28 cells total, 17 show $M<1$, range $[0.81, 1.00]$, mean $0.96$. Minimum at $(\theta{=}0.95, \text{slice}{=}0.10)$, the most aggressive cell tested, at $0.81$. This monotone degradation rules out the ``$M{=}1$ is trivial'' concern empirically rather than rhetorically.

\section{Measured Compute \& Energy (Extended)}\label{app:realcompute}

Per-run instrumentation: \texttt{nvidia-smi --query-gpu=power.draw} sampled at eval start and end; integrated with wall-clock elapsed. $1\times$H200 (CUDA\_VISIBLE\_DEVICES isolated per shard).

\begin{figure}[h]
\centering
\includegraphics[width=0.75\textwidth]{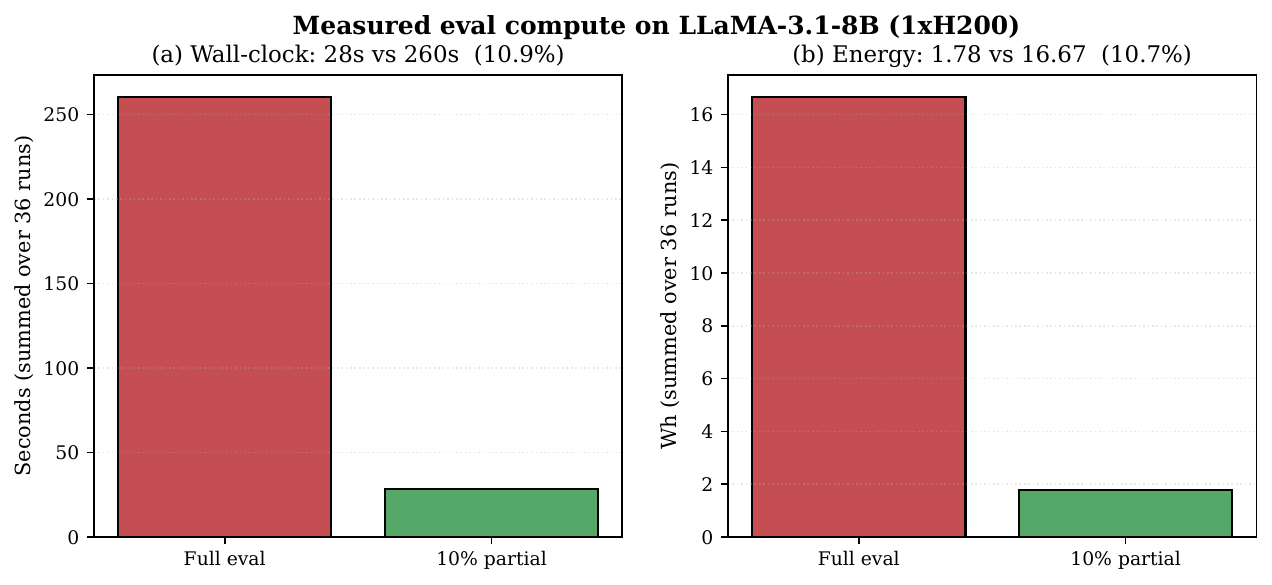}
\caption{\textbf{Measured evaluation compute on LLaMA-3.1-8B (1$\times$H200).} Summed over 36 trained runs: $10\%$ slice evaluation uses $10.9\%$ of full-eval wall-clock and $10.7\%$ of full-eval energy---an $89\%$ reduction in both, with ratios agreeing to $0.2\%$.}
\label{fig:realcompute}
\end{figure}

\textbf{Per-cycle cost.} Full cycle: $C_{\text{cycle}}^{\text{naive}} = C_{\text{label}} + C_{\text{train}} + C_{\text{eval}}^{\text{full}}$. Under the protocol at steady state: $C_{\text{cycle}}^{\text{TG}} = C_{\text{label}} + C_{\text{train}} + (\alpha + 1/N) \cdot C_{\text{eval}}^{\text{full}}$. For $\alpha{=}0.10$, $N{=}10$: $C_{\text{eval}}^{\text{TG}} = 0.20 \cdot C_{\text{eval}}^{\text{full}}$, an $80\%$ evaluation-stage saving (consistent with the $66\%$ trajectory figure, which includes calibration overhead).

\textbf{Annualized scale.} Per single-candidate evaluation cycle, the savings are $\approx 6.4$ seconds wall-clock and $\approx 0.41$ Wh energy (Table~\ref{tab:realcompute} aggregates summed over 36 runs and divided to per-run units). At daily cadence ($365$ cycles/year) on $10$ model families with one candidate per cycle, this extrapolates to $\approx 6.5$ GPU-hours/year and $\approx 1.5$ kWh/year per model family---small in absolute terms but representative of the per-candidate efficiency gain. At $36$-candidate batch cadence (matching our experimental sweep cadence), the same arithmetic yields $\approx 235$ GPU-hours/year and $\approx 54$ kWh/year.

\textbf{Caveats.} H100/H200 figures are specific to our hardware; relative savings ($89\%$) transfer more robustly than absolute numbers. I/O-bound evaluations (e.g., large video retrieval) will show smaller ratios; our compute-bound regime is representative of standard classification benchmarks.

\section{Multi-Metric Gates (Adult + Fairness)}\label{app:multimetric}

We extend $M$ to a joint gate $G_{\text{multi}} = (F_1 \ge \tau_{F_1}) \wedge (\text{DP-diff} \le \tau_{\text{DP}})$ on Adult with \texttt{sex} as protected attribute. XGBoost, 5 seeds $\{7,8,9,42,123\}$.

\begin{table}[h]
\caption{Multi-metric $M$ on Adult. Seed=123 sits at the $F_1$ boundary and triggers fallback under both gates.}
\label{tab:multimetric}
\begin{center}\begin{small}\begin{sc}
\begin{tabular}{lcccc}
\toprule
setting & $\tau_{F_1}$ & $\tau_{\text{DP}}$ & single-$M$ & multi-$M$ \\
\midrule
conservative & $0.50$ & $0.30$ & $1.00$ & $1.00$ \\
tight        & $0.70$ & $0.15$ & $0.80$ & $0.80$ \\
\bottomrule
\end{tabular}
\end{sc}\end{small}\end{center}
\end{table}

Under conservative thresholds both gates hold at $M{=}1$; under tight thresholds both drop to $0.80$. The single failing seed (\texttt{seed{=}123}, full-eval $F_1{=}0.716$, DP-diff$=0.189$) sits at the $F_1$ boundary and fails both gate types identically. \textbf{Interpretation}: $M$ extends to multi-metric gates without loss of reliability; the combined gate does not artificially inflate $M{=}1$, and it correctly triggers fallback when the model sits near the joint threshold surface.

\begin{figure}[h]
\centering
\includegraphics[width=0.6\textwidth]{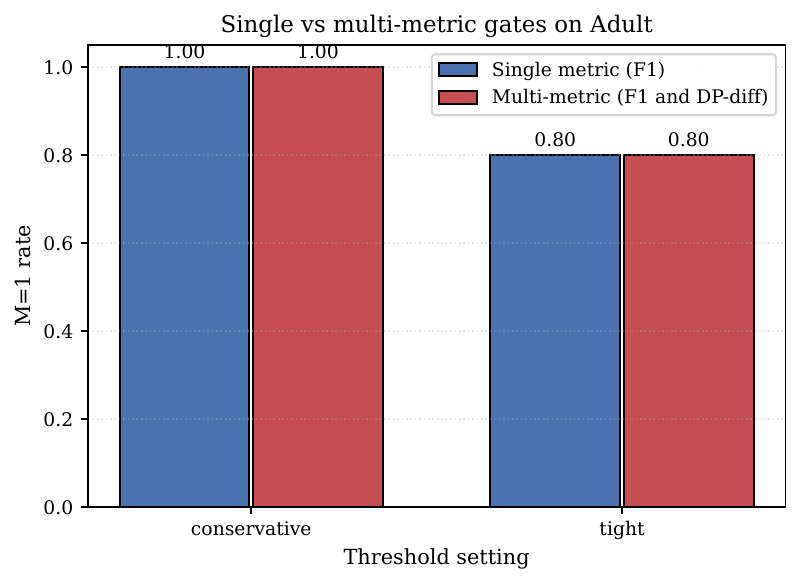}
\caption{Multi-metric gate extension on Adult. Both single-metric and multi-metric $M$ drop from $1.00$ to $0.80$ as thresholds tighten.}
\label{fig:multimetric}
\end{figure}

\section{Sentinel Protocol: Full Trajectory Results}\label{app:sentinel}

100-cycle continual-adaptation simulation, drawing from the LLaMA run pool, with $K{=}10$ calibration cycles, $\epsilon{=}0.02$, partial slice $10\%$:

\begin{table}[h]
\caption{Protocol savings vs. sentinel audit period $N$ (100 cycles, threshold $0.80$).}
\label{tab:sentinel}
\begin{center}\begin{small}\begin{sc}
\begin{tabular}{ccccc}
\toprule
$N$ & compute (TG) & compute (naive) & savings & fallbacks \\
\midrule
$5$   & $43.0$  & $100$ & $57\%$ & $5$ \\
$10$  & $34.0$  & $100$ & $\mathbf{66\%}$ & $5$ \\
$20$  & $29.0$  & $100$ & $71\%$ & $5$ \\
$50$  & $26.0$  & $100$ & $74\%$ & $5$ \\
$100$ & $25.0$  & $100$ & $75\%$ & $5$ \\
\bottomrule
\end{tabular}
\end{sc}\end{small}\end{center}
\end{table}

\begin{figure}[h]
\centering
\includegraphics[width=0.75\textwidth]{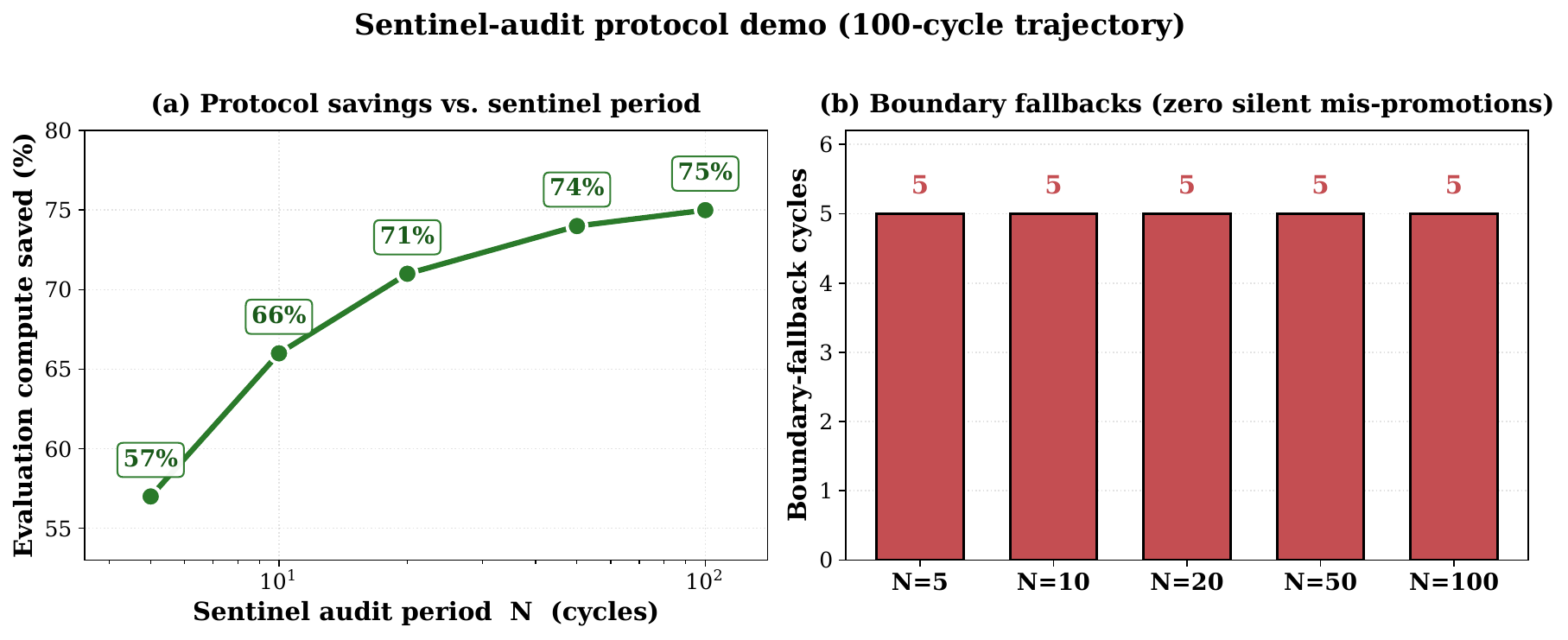}
\caption{\textbf{100-cycle continual-adaptation simulation under the operational protocol.} Savings rise from $57\%$ ($N{=}5$) to $75\%$ ($N{=}100$); headline: $N{=}10$, $66\%$. Every configuration has $5$ boundary-fallback cycles---\textbf{zero silent mis-promotions in this simulated trajectory}.}
\label{fig:sentinel}
\end{figure}

Calibration $M{=}1$ rate: $1.00$ across all $K{=}10$ cycles (consistent with the $35/36$ LLaMA result). Silent mis-promotions: $0$ across all configurations. The constant 5 boundary fallbacks reflect the same near-threshold runs catching the protocol's $\epsilon$-margin check---exactly the intended behavior.

\textbf{Reading these numbers under distribution shift.} The pool is drawn from a fixed SST-2 distribution, so silent mis-promotions are zero \emph{by construction of the pool} rather than by guarantee of the protocol. To interpret the table for shift-prone deployments: read the sentinel period $N$ as the worst-case detection lag (cycles between a shift-induced disagreement and its observation), and the boundary-fallback count as the protocol's intra-cycle safety margin. Smaller $N$ shortens detection latency at the cost of savings (the $N{=}5$ row gives the lower bound on savings in our trajectory at $57\%$); rolling recalibration that reverts to dual evaluation when the trailing agreement rate slips below target is the principled extension and the subject of follow-up work (\cref{app:limitations}).

\section{Limitations and Future Work}\label{app:limitations}

\textbf{Scope.} We tested two model classes (XGBoost, LLaMA-3.1-8B) and one domain each. Extension to vision, multimodal, and agentic settings is ongoing; the mechanism does not require retraining to extend.

\textbf{Threshold sensitivity.} Our sensitivity sweep establishes $M$ is informative at tight thresholds, but in new domains calibration data is required to set $\alpha, N, \epsilon$. A minimum $K{\ge}5$ calibration cycles is recommended before deploying partial-only steady state.

\textbf{Multi-metric calibration.} The Adult demonstration shows $M$ extends to multi-metric gates; scaling to production bundles (quality $\wedge$ fairness $\wedge$ latency $\wedge$ cost) requires per-metric threshold calibration we leave to future work. Sequential-testing-based boundary detection is a natural refinement of the $\epsilon$-margin rule.

\textbf{Cost model.} $c(u)$ assumes homogeneous hardware. Heterogeneous clusters require per-node-class estimation; integrating production telemetry is planned.

\textbf{Dynamics over many cycles.} We conducted simulations for up to 100 cycles. Long-horizon dynamics, such as whether label-first remains optimal as models evolve and when the Pareto shift occurs, necessitate longitudinal studies.

\textbf{Distribution shift.} The most consequential limitation is that all of our empirical pool is drawn under SST-2's fixed distribution (and Adult's), so the 100-cycle simulation characterizes \emph{protocol mechanics} (calibration $\to$ partial-only $\to$ sentinel $\to$ boundary fallback) rather than \emph{shift detection}. Concretely: if a shift occurs shortly after the calibration phase and causes partial-eval decisions to silently diverge from full-eval decisions, the protocol's worst-case detection latency is one sentinel period $N$. The boundary-fallback margin $\epsilon$ partially mitigates this by forcing full evaluation at exactly the cycles where shift is most likely to flip a decision (those near threshold). For shift-sensitive deployments we recommend: (i) small $N$ (we report $N{=}5$ in \cref{app:sentinel} at $57\%$ savings as the lower bound on savings in our trajectory); (ii) wider $\epsilon$ to widen the safety band around the threshold; (iii) \emph{rolling recalibration} --- maintain a trailing estimate $\widehat{P}(M{=}1)$ on the most recent $W$ cycles and revert to dual evaluation whenever the estimate falls below target. We do not validate (iii) empirically here; characterizing $M$ under controlled covariate shift (e.g., GLUE-X, WILDS-style shift benchmarks, or synthetic drift overlaid on SST-2) is the most important follow-up for production adoption.

\textbf{Dataset representativeness.} SST-2 is a relatively saturated benchmark for LLaMA-3.1-8B, which contributes to the high $M{=}1$ rate (35/36) at the conservative $0.80$ threshold. On more heterogeneous tasks --- where full-eval scores land closer to the threshold, or where class imbalance interacts with random slicing --- we expect the $M$ rate to be lower at conservative thresholds, and we expect adaptive (non-random) subset selection to help. The sensitivity sweep in \cref{app:sensitivity} simulates this regime by tightening the threshold rather than changing the dataset; a multi-dataset characterization (e.g., GLUE, HELM subsets, BIG-Bench-Lite) is required to estimate the cross-task $M$ distribution and is part of our planned follow-up.

\textbf{What the $2.3\times$ labeling-vs-training ratio is and is not.} The $2.3\times$ marginal-gain figure is computed under a fixed-budget allocation sweep in which ``labeling'' reveals pre-existing labels at simulated rate $\lambda$ and ``training'' is additional epochs on the resulting larger labeled pool. It is therefore best read as \emph{the marginal value of an additional labeled sample versus an additional training epoch, holding total cycle compute constant.} It is \emph{not} a claim that an annotation workflow and a training workflow can be freely swapped within one short cycle: real annotation pipelines involve upstream procurement, quality control, and human-in-loop latency that we do not model, and that typically operate on a longer horizon than a single cycle. The result is a useful guide to \emph{budget allocation across stages a team can already flex} (e.g., which queued labels to release, how many epochs to schedule), not to upstream annotation planning.

\textbf{Eval-stage scope of empirical savings.} Although the framework formally budgets labeling, training, and evaluation jointly, the compute savings we measure are concentrated in evaluation. Reallocating saved evaluation wall-clock toward labeling or training within the same cycle is a deployment-time choice that depends on cluster scheduling and pipeline topology, and we intentionally do not specify it inside \textsc{TimeGate} --- separating measurement (which the policy layer provides) from reallocation (which depends on production constraints) keeps the layer thin. The corollary is that ``$66\%$ savings'' should be read as ``$66\%$ of the evaluation stage'' rather than ``$66\%$ of the cycle''; cycle-level savings depend on the ratio $C_{\text{eval}}^{\text{full}} / C_{\text{cycle}}^{\text{naive}}$, which is workload-specific.

\textbf{Future Scope of work.} Several avenues extend beyond the scope of this paper and form the basis of our primary follow-up work: (i) conducting a systematic comparison with sequential-testing and confidence-interval baselines, as well as a composed comparison with adaptive-subset selection methods such as SubLIME~\citep{xuDataEfficientEvaluation2024} and SimBA~\citep{leeSimBaSimplicityBias2025} (substituting their selectors for the random slice and re-auditing through $M$); (ii) expanding to multi-task foundation-model benchmarks, including MMLU and HELM subsets, and incorporating at least one additional open FM family; (iii) performing longitudinal studies on 500+ cycle drift-aware benchmarks with controlled covariate shift, and validating the rolling-recalibration extension introduced above; (iv) conducting ablations on calibration length $K$, sentinel period $N$, boundary margin $\epsilon$, and slice size $\alpha$ under realistic drift conditions; and (v) extending energy/time measurements across various hardware types and workloads to confirm that the $\sim$89\% savings ratio is applicable beyond single-H200 classification. We consider these as natural progressions rather than essential for establishing the workshop-scoped claims of the current paper.

\section{Impact Statement}\label{app:impact}

This paper develops a policy layer for governing continual ML adaptation
under jointly budgeted time, labeling, training, and evaluation resources.
We discuss the broader impacts we judge most relevant.

\textbf{Sustainability and energy.} Continual adaptation pipelines are a
recurring source of compute and energy expenditure across deployed ML
systems. The mechanism we describe reduces evaluation compute and energy
substantially per cycle on the workloads we measure ($\sim$89\% reduction
in wall-clock and energy on a single H200 for the LLaMA-3.1-8B
configuration). To the extent partial-evaluation calibration generalizes
to other production pipelines, the aggregate effect on data-center energy
demand and associated carbon footprint of continual ML systems is
positive, though the absolute magnitude depends strongly on deployment
scale and hardware mix, which we do not measure here. Conversely, savings
that lower the marginal cost of continual adaptation could induce more
adaptation cycles per system (a Jevons-paradox concern stated in \citet{luccioniEfficiencyGainsRebound2025}), partially
offsetting per-cycle gains; we believe this is a worthwhile trade given
that more frequent adaptation also enables better drift response and
distributional fairness over time, but it warrants attention in production
deployments.

\textbf{Reliability and silent failure.} The metric-availability signal
$M$ is designed to make partial-evaluation deployment auditable:
calibration cycles surface when partial decisions disagree with full
decisions, and a boundary-fallback mechanism forces full evaluation near
decision boundaries. We are explicit in the paper that $M$ is asymmetric
(false positives are costlier than false negatives) and that our
zero-mis-promotion result is measured in a simulated trajectory drawn
from observed agreement distributions, not under independent drift.
Production teams adopting this protocol should treat the calibration
phase as load-bearing and should not deploy partial-only steady state
without first establishing a domain-specific empirical agreement rate.

\textbf{Fairness and multi-metric decisions.} Promotion gates in
production typically bundle quality, fairness, and cost criteria. Our
multi-metric demonstration on Adult ($F_1 \wedge$ demographic-parity-diff)
shows the mechanism extends to bundled gates without loss of reliability
under our setup, but reliable fairness gating in real deployments
requires per-metric threshold calibration, attention to subgroup
representation in the partial slice, and evaluation of how slicing
interacts with rare protected groups. We have not characterized this
interaction systematically and view it as essential follow-up work.

\textbf{Human labor.} The paper treats labeling time as a budgeted
resource. Our experimental simulations assume idealized annotator
throughput; deployments involving human annotators should consider
working conditions, fair compensation, and the well-documented downstream
effects of annotation pipelines on labor practices. Our framework is
neutral to these choices but does not address them.

\textbf{Misuse and dual use.} The mechanism we describe is a release-gate
policy layer; it does not generate new model capabilities and we do not
foresee specific dual-use risks beyond those already inherent to the
underlying continually-adapted models, which the user community is
actively studying.

\end{document}